\documentclass[11pt,a4paper]{article}

\usepackage[utf8]{inputenc}
\usepackage[T1]{fontenc}
\usepackage{lmodern}
\usepackage[margin=1in]{geometry}
\usepackage{graphicx}
\usepackage{booktabs}
\usepackage{array}
\usepackage{multirow}
\usepackage{longtable}
\usepackage{hyperref}
\usepackage{xcolor}
\usepackage{amsmath}
\usepackage{amssymb}
\usepackage{natbib}
\usepackage{setspace}
\usepackage{lineno}
\usepackage{float}
\usepackage{caption}
\usepackage{subcaption}
\usepackage{enumitem}
\usepackage{textcomp}

\hypersetup{
    colorlinks=true,
    linkcolor=blue,
    filecolor=magenta,
    urlcolor=cyan,
    citecolor=blue,
    pdftitle={Benchmarking an Evidence-Grounded Clinical Intelligence Layer on the ESAP 2025 Endocrinology Examination},
    pdfauthor={Amir Hosseinian et al.},
}

\newcommand{\Mirror}{\textsc{Mirror}}

\begin{document}

\begin{center}
{\LARGE \textbf{Benchmarking an Evidence-Grounded Clinical Intelligence Layer on the ESAP 2025 Endocrinology Examination}}

\vspace{1em}

{\large
Amir Hosseinian\textsuperscript{1},
Reza Shahneh\textsuperscript{1},
Umer Mansoor\textsuperscript{1},
Gilbert Szeto\textsuperscript{1},
Kirill Karlin\textsuperscript{1},
Nima Aghaeepour\textsuperscript{2}
}

\vspace{0.5em}

{\small
\textsuperscript{1}January AI; \textsuperscript{2}Stanford University
}

\vspace{0.5em}

{\small \textit{Corresponding author:} \href{mailto:amirhoss@january.ai}{amirhoss@january.ai}}

\end{center}

\vspace{1em}

\begin{abstract}
\noindent\textbf{Background:} Large language models have demonstrated strong performance on general medical examinations, but subspecialty clinical reasoning remains challenging due to rapidly evolving guidelines and nuanced evidence hierarchies.

\noindent\textbf{Methods:} We evaluated January \Mirror{}, an evidence-grounded clinical reasoning system, against frontier LLMs (GPT-5, GPT-5.2, Gemini-3-Pro) on the Endocrine Self-Assessment Program (ESAP) 2025, a 120-question board-style examination. \Mirror{} integrates a curated endocrinology and cardiometabolic evidence corpus with a structured reasoning architecture to generate evidence-linked outputs. \Mirror{} operated under a closed-evidence constraint without external retrieval. Comparator LLMs had real-time web access to guidelines and primary literature.

\noindent\textbf{Results:} \Mirror{} achieved 87.5\% accuracy (105/120; 95\% CI: 80.4--92.3\%), exceeding the human reference of 62.3\% (ESAP respondent distributions) and frontier LLMs including GPT-5.2 (74.6\%), GPT-5 (74.0\%), and Gemini-3-Pro (69.8\%). On the 30 most difficult questions (human accuracy $<50$\%), \Mirror{} achieved 76.7\% accuracy. Top-2 accuracy was 92.5\% for \Mirror{} versus 85.25\% for GPT-5.2.

\noindent\textbf{Conclusions:} \Mirror{} provided evidence traceability: 74.2\% of outputs cited at least one guideline-tier source, with 100\% citation accuracy on manual verification. Curated evidence with explicit provenance can outperform unconstrained web retrieval for subspecialty clinical reasoning and supports auditability for clinical deployment.
\end{abstract}

\vspace{0.5em}

\vspace{1em}

\section{Introduction}

The successful performance of large language models on the United States Medical Licensing Examination generated significant enthusiasm about AI's potential in clinical medicine. In 2023, Kung et al.\ demonstrated that ChatGPT performed at or near the passing threshold for all three USMLE steps without specialized training \citep{kung2023}, and subsequent work by Nori et al.\ showed that GPT-4 exceeded the USMLE passing score by over 20 points \citep{nori2023}. However, the USMLE assesses general medical knowledge expected of all physicians, not the specialized reasoning required for subspecialty practice. Real-world clinical decisions frequently demand subspecialty expertise, the ability to integrate current society guidelines, interpret nuanced clinical evidence, and reason across complex patient presentations with multiple comorbidities.

Endocrinology presents a particularly demanding test case. The American Board of Internal Medicine (ABIM) reports that endocrinology certification pass rates have shown notable variability, falling to 74\% for first-time takers in recent years before recovering to 85\% in 2024 \citep{abim2024}. This difficulty stems from several factors: the rapid evolution of treatment guidelines (particularly for diabetes, obesity, and thyroid disorders), the complex interplay between metabolic systems, and the need to integrate emerging therapeutic classes such as GLP-1 receptor agonists and SGLT2 inhibitors into established treatment algorithms. A system capable of reliable endocrinology reasoning would demonstrate capabilities directly relevant to clinical practice.

Current approaches to improving LLM clinical performance have focused primarily on model scale and retrieval augmentation. Larger models with more parameters, combined with real-time web search capabilities, represent the dominant paradigm. Yet this approach faces fundamental limitations: web retrieval returns heterogeneous sources without consistent evidence grading, provenance normalization, or domain-scoped filtering, which can reduce reliability even when the underlying information is available \citep{kohandel2025}. There is no mechanism for weighting evidence by source quality, recency, or clinical relevance. Furthermore, LLMs are known to produce hallucinations, factually incorrect or fabricated information, which pose particular risks in clinical settings where errors can compromise patient safety \citep{asgari2025,roustan2025}.

While retrieval augmentation improves factual recall, it does not impose constraints on evidence hierarchy, guideline precedence, or clinical relevance. As a result, models may retrieve correct information yet fail to apply it appropriately in context.

We hypothesized that an alternative approach, domain-specific evidence curation combined with structured clinical reasoning, could outperform model scale and general web retrieval. To test this hypothesis, we developed \Mirror{}, an evidence-grounded clinical reasoning system built around a curated endocrinology and cardiometabolic evidence corpus and a structured clinical reasoning stack designed to produce evidence-linked outputs. We evaluated \Mirror{} against frontier LLMs on the Endocrine Self-Assessment Program (ESAP) 2025, a rigorous board-style examination developed by the Endocrine Society for certification maintenance \citep{esap2025}.

\Mirror{} is a broader clinical platform; however, this paper evaluates only the evidence-grounded clinical intelligence layer responsible for producing evidence-linked, verifiable outputs. We focus on this layer because credibility and traceability are prerequisites for clinician-facing and enterprise clinical workflows, independent of how patient context is sourced or represented.

This paper is the first in a planned validation series focused on the credibility of \Mirror{}'s clinical intelligence layer. This study benchmarks subspecialty exam performance; the next paper extends evaluation across additional standardized examinations; and the third paper evaluates verifiability and clinical usefulness via blinded clinician review on complex case vignettes. Together, the series is intended to establish whether \Mirror{}'s outputs are reliable enough to support clinician-facing and enterprise clinical workflows.

\section{Methods}

\subsection{Benchmark Dataset}

We utilized the Endocrine Self-Assessment Program (ESAP) 2025, developed by the Endocrine Society as a self-assessment tool for endocrinologists preparing for board certification and maintenance of certification examinations \citep{esap2025}. ESAP consists of 120 multiple-choice questions spanning the breadth of clinical endocrinology, including disorders of the thyroid, adrenal, pituitary, parathyroid, and reproductive systems, as well as diabetes mellitus, lipid disorders, and obesity medicine. Each question presents a clinical vignette with patient history, physical examination findings, and laboratory results, followed by a single-best-answer format.

\begin{figure}[htbp]
\centering
\includegraphics[width=0.9\textwidth]{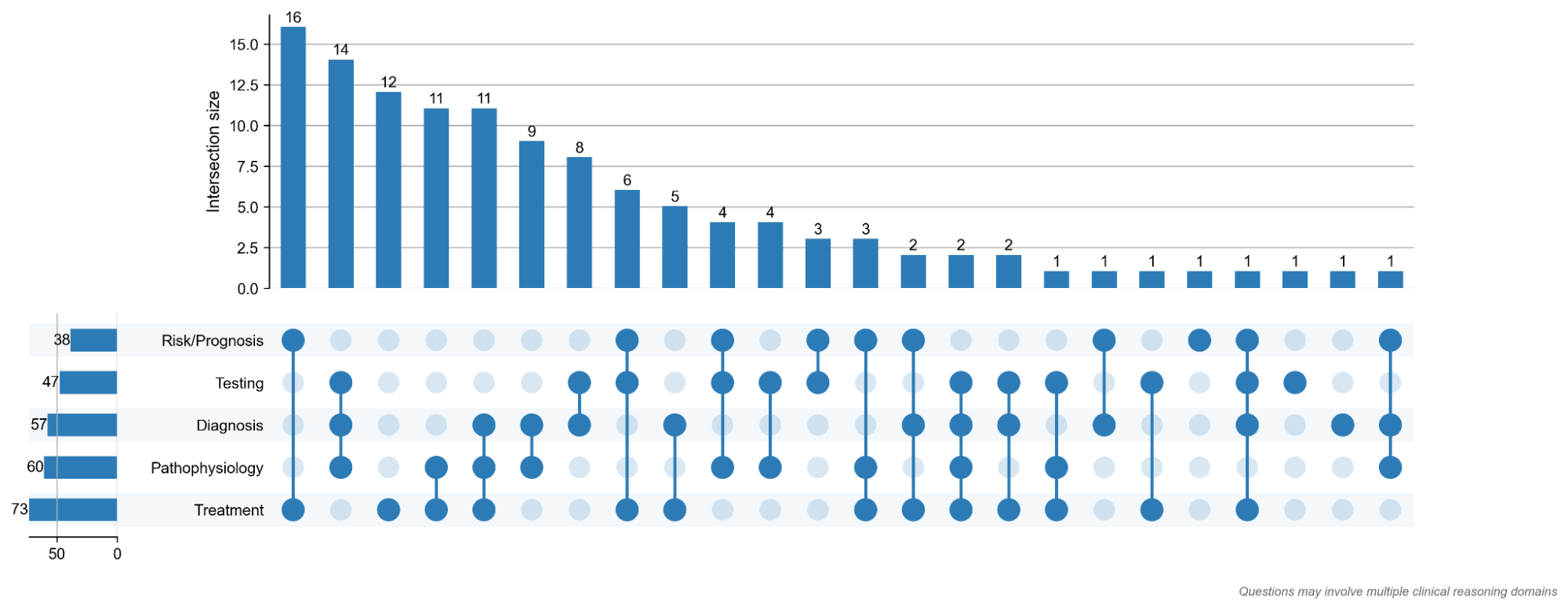}
\caption{Distribution of clinical reasoning domains in ESAP 2025. The left panel shows the total number of questions involving each domain. The main panel displays intersection sizes for domain combinations, sorted by frequency. The matrix indicates which domains comprise each intersection. Treatment-related reasoning was most prevalent (n=73), followed by pathophysiology (n=60) and diagnosis (n=57). The most common domain combination was risk/prognosis with treatment (n=16).}
\label{fig:domains}
\end{figure}

Questions were categorized by clinical reasoning type, with each question potentially involving multiple reasoning domains: diagnosis (47.5\% of questions), pathophysiology (50.0\%), treatment selection (60.8\%), diagnostic testing (39.2\%), and risk assessment/prognosis (31.7\%). This multi-label categorization reflects the integrated nature of clinical reasoning, where a single case may require diagnostic, therapeutic, and prognostic considerations. The examination is designed to assess reasoning at the level expected of a practicing endocrinologist and meets ABIM standards for maintenance of certification. Figure~\ref{fig:domains} displays the distribution of clinical reasoning domains and their co-occurrence patterns across the examination.

\subsection{Systems Evaluated}

We compared \Mirror{} against three frontier LLMs: GPT-5, GPT-5.2, and Gemini-3-Pro. Frontier LLMs were evaluated with unrestricted web access enabled, allowing each model to perform as many web queries as needed to answer each question. \Mirror{} operated without external web access and relied exclusively on its curated evidence base. This design isolates the contribution of domain-curated evidence and structured reasoning relative to general-purpose models augmented with web retrieval.

\begin{table}[htbp]
\centering
\caption{Systems evaluated and their capabilities}
\label{tab:systems}
\begin{tabular}{@{}lcll@{}}
\toprule
\textbf{System} & \textbf{Web Access} & \textbf{Evidence Base} & \textbf{Architecture} \\
\midrule
\Mirror{} & No & Curated endocrinology corpus & Evidence-grounded ensemble \\
GPT-5.2 & Yes & Real-time web retrieval & Monolithic LLM \\
GPT-5 & Yes & Real-time web retrieval & Monolithic LLM \\
Gemini-3-Pro & Yes & Real-time web retrieval & Monolithic LLM \\
\bottomrule
\end{tabular}
\end{table}

\subsection{Web-Assisted Baseline Protocol}

For web-assisted baselines, browsing was enabled with unrestricted access and consistent instructions to prioritize primary society guidelines, peer-reviewed articles, and authoritative medical references. Retrieved content was provided to the model as context for the final answer within the same interaction, and the model was required to output a single best answer choice.

\subsection{\Mirror{} System Architecture}

\Mirror{} is designed as a clinical reasoning system that grounds outputs in curated medical evidence. The system comprises two primary components: a domain-specific evidence corpus and a structured reasoning architecture.

\textbf{Evidence Corpus.} \Mirror{}'s evidence corpus is organized in a multi-tiered hierarchy reflecting clinical evidence standards in endocrinology and cardiometabolic medicine. The first tier comprises society-level guidelines and practice statements from authoritative bodies including the Endocrine Society, American Diabetes Association (ADA), American Association of Clinical Endocrinology (AACE), Kidney Disease: Improving Global Outcomes (KDIGO), and related cardiometabolic guideline organizations \citep{ada2024,acc2019,kdigo2022,aace2023,aasld2023}. The second tier includes high-impact peer-reviewed clinical literature from journals with established methodological rigor relevant to endocrinology and internal medicine. The third tier comprises landmark and pivotal late-phase randomized controlled trials that have shaped current practice, particularly for therapeutic classes such as GLP-1 receptor agonists, dual incretin agonists, SGLT2 inhibitors, and obesity interventions.

\textbf{Reasoning Architecture.} \Mirror{} uses an ensemble-style clinical reasoning stack in which multiple specialized reasoning components generate candidate answers and supporting evidence links, followed by an arbitration layer that selects the final output using evidence quality and internal agreement signals. The components are organized around common clinical question archetypes (e.g., diagnosis, testing, treatment, prognosis, and mechanistic reasoning) to reflect how clinicians approach distinct reasoning tasks. To avoid disclosing proprietary implementation details, we report the system's inputs/outputs and evaluation behavior while omitting internal ranking heuristics and model-specific routing parameters.

\subsection{Evaluation Protocol}

All systems were evaluated under zero-shot conditions without access to ESAP questions or answers during development or training. Questions were presented without preprocessing, exactly as they appear in the examination.

The primary outcome measure was single-answer accuracy: the proportion of questions for which the system selected the correct answer. As a secondary outcome, we evaluated top-2 accuracy, whether the correct answer appeared among the system's top two selections. This metric reflects a clinical decision support workflow in which physicians use AI to narrow options rather than provide definitive single answers.

Performance was stratified by question type (diagnosis, testing, treatment, risk/prognosis, pathophysiology) to identify domain-specific strengths and weaknesses. Statistical comparisons between systems used McNemar's test for paired proportions, with significance set at $p < 0.05$. Confidence intervals (95\%) were calculated using the Wilson score method.

\subsection{Citation Verification Protocol}

To assess evidence traceability and citation accuracy, we conducted a manual verification audit of \Mirror{}'s outputs. For each of the 120 questions, \Mirror{} produced an answer accompanied by citations to specific evidence sources from its curated corpus. A random sample of 60 questions was selected for detailed citation verification.

For each sampled question, a clinical reviewer assessed: (1) whether the cited source existed in the corpus, (2) whether the cited passage was accurately quoted or paraphrased, and (3) whether the cited evidence logically supported the selected answer. Citations were classified as ``verified accurate'' if all three criteria were met.

Additionally, we assessed guideline concordance for treatment-related questions. For each treatment question ($n = 73$), the relevant society guideline recommendation was identified, and the system's answer was scored as guideline-concordant or guideline-discordant.

\section{Results}

\subsection{Overall Performance}

\Mirror{} substantially outperformed all frontier LLMs despite operating without web access. \Mirror{} achieved 87.5\% accuracy on the ESAP 2025 examination ($n = 120$ questions; 95\% CI: 80.4--92.3\%), compared to 74.6\% for GPT-5.2 (95\% CI: 66.7--82.5\%), 74.0\% for GPT-5 (95\% CI: 66.7--80.0\%), and 69.8\% for Gemini-3-Pro (95\% CI: 61.6--77.1\%), all of which had real-time web access to clinical resources. Pairwise McNemar's tests confirmed statistically significant differences between \Mirror{} and all three baselines: GPT-5.2 ($p = 0.011$), GPT-5 ($p < 0.001$), and Gemini-3-Pro ($p < 0.001$). All comparisons remained significant after Bonferroni correction for multiple testing ($\alpha = 0.017$). While confidence intervals overlap between some systems, the paired analysis demonstrates consistent directional advantage for \Mirror{}.

\begin{figure}[htbp]
\centering
\includegraphics[width=0.9\textwidth]{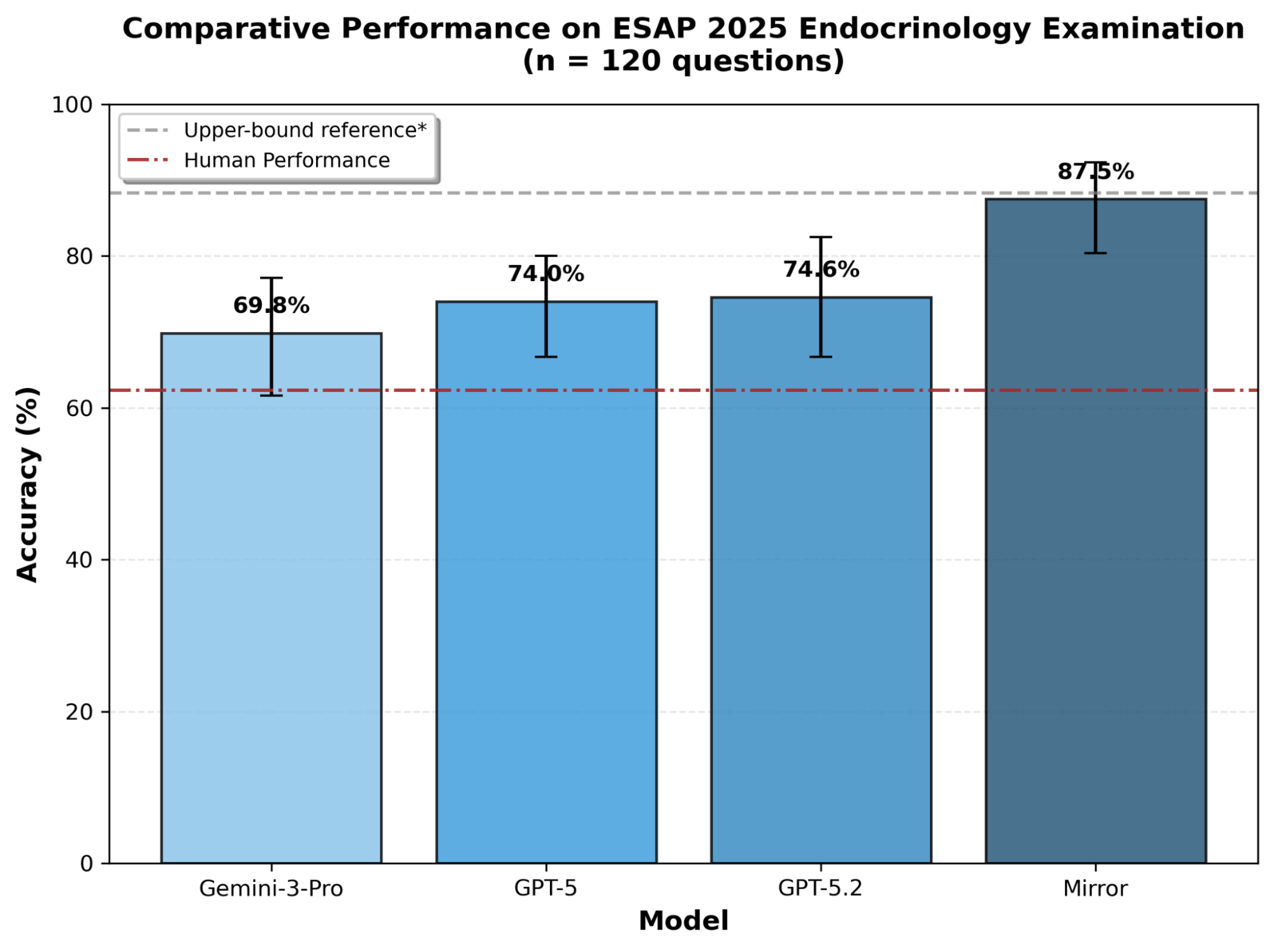}
\caption{Comparative performance on ESAP 2025 endocrinology examination. \Mirror{} achieved 87.5\% accuracy compared to 74.6\% for the best-performing frontier LLM (GPT-5.2) and 62.3\% for the human reference (ESAP respondent mean). Optional dashed line indicates an internal upper-bound reference condition, provided for context. Error bars represent 95\% confidence intervals. *$p < 0.05$ vs.\ \Mirror{}.}
\label{fig:performance}
\end{figure}

Figure~\ref{fig:performance} displays the comparative performance across all systems, including the ESAP respondent reference derived from aggregate answer distributions. \Mirror{}'s performance approached an internal upper-bound reference condition in which a baseline model was provided with manually selected, question-specific evidence passages.

\subsection{Performance Relative to ESAP Respondent Reference}

ESAP provides aggregate respondent statistics indicating the percentage of test-takers who selected each answer option. We used these distributions to calculate an item-level respondent correctness rate of 62.3\% (SD = 16.8\%), representing the mean percentage of ESAP respondents selecting the correct answer across all 120 questions. This metric serves as a proxy for human performance, though ESAP respondents are self-selected endocrinologists who may have used external resources during self-assessment. Individual question difficulty ranged from 22.3\% to 94.4\% correct.

\begin{figure}[htbp]
\centering
\includegraphics[width=0.9\textwidth]{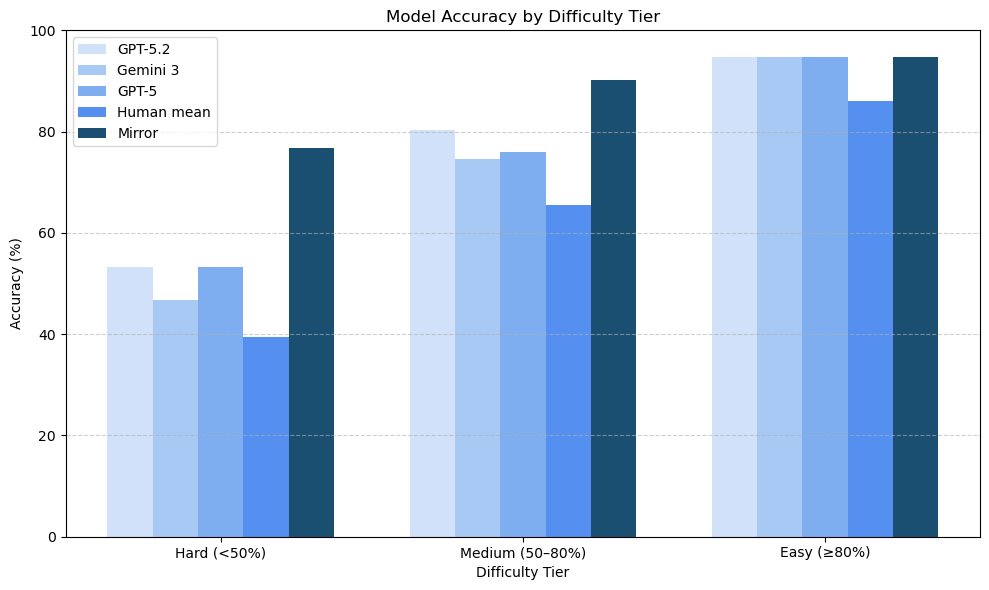}
\caption{Performance stratified by question difficulty (based on human accuracy). \Mirror{} maintained strong performance across all difficulty tiers, with particular advantage on hard questions (human accuracy $<$50\%) where it achieved 76.7\% compared to 53.3\% for GPT-5.2 and 39.5\% human mean accuracy.}
\label{fig:difficulty}
\end{figure}

\Mirror{}'s 87.5\% accuracy exceeded this respondent reference by 25.2 percentage points. To assess whether this advantage extended to the most challenging clinical scenarios, we stratified questions by human performance. On the 30 most difficult questions (human accuracy $<$50\%, mean 39.5\%), \Mirror{} achieved 76.7\% accuracy (23/30), compared to 53.3\% for GPT-5.2, 53.3\% for GPT-5, and 46.7\% for Gemini-3-Pro. On 71 medium-difficulty questions (human accuracy 50--80\%, mean 65.6\%), \Mirror{} achieved 90.1\% versus 80.3\% for GPT-5.2. On 19 easy questions (human accuracy $\geq$80\%), all systems performed well (\Mirror{} 94.7\%, GPT-5.2 94.7\%). Figure~\ref{fig:difficulty} illustrates performance stratified by question difficulty.

\begin{figure}[htbp]
\centering
\includegraphics[width=0.9\textwidth]{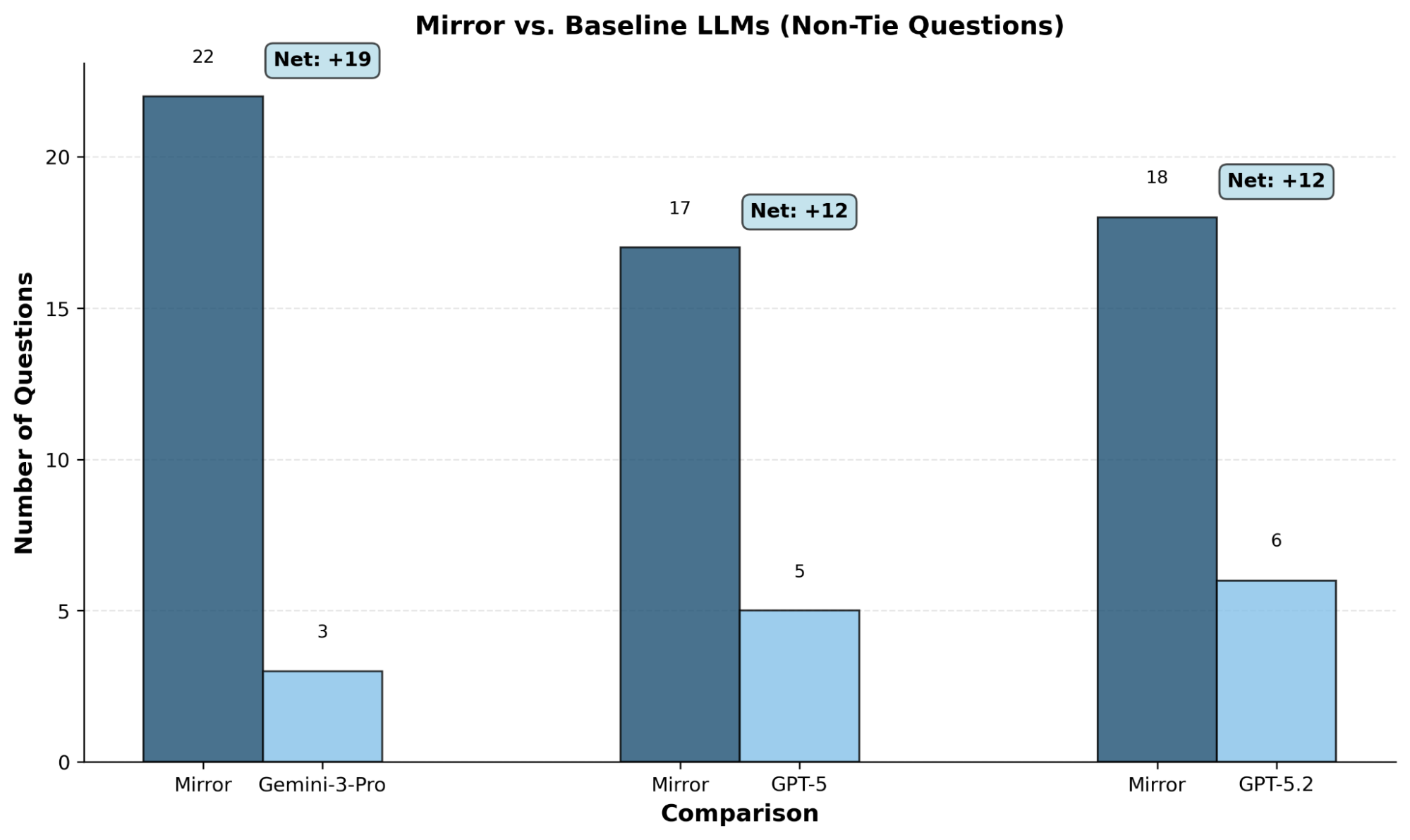}
\caption{Paired win/loss analysis comparing \Mirror{} to each baseline. For each comparison, ``wins'' indicate questions \Mirror{} answered correctly while the baseline erred; ``losses'' indicate the reverse. \Mirror{} achieved a net positive margin against all baselines, with the largest advantage over Gemini-3-Pro (19 net wins).}
\label{fig:winloss}
\end{figure}

\subsection{Top-2 Accuracy}

When evaluating top-2 selections, \Mirror{} achieved 92.5\% accuracy (111/120 questions; 95\% CI: 86.4--96.0\%), indicating that the correct answer appeared among \Mirror{}'s two highest-ranked options in 92.5\% of questions. This metric is clinically relevant because decision support systems are often used to narrow differential diagnoses or treatment options rather than provide single definitive answers. Comparable top-2 accuracy figures for frontier LLMs were 85.25\% (GPT-5.2), 81.1\% (GPT-5), and 83.59\% (Gemini-3-Pro). The top-2 accuracy gap (7.3 percentage points vs.\ GPT-5.2) indicates that frontier LLMs more frequently excluded the correct answer from their top two selections, reducing their utility as differential-narrowing tools in clinical workflows.

\subsection{Performance by Question Type}

Table~\ref{tab:accuracy} presents accuracy stratified by clinical question type. Note that questions may involve multiple reasoning domains; the counts reflect all questions tagged with each category. \Mirror{} achieved its highest accuracy on diagnosis questions (91.2\%) and demonstrated consistent performance across all categories (85.1--91.2\%). The performance differential with frontier LLMs was largest in treatment questions, with a 15.5 percentage point difference compared to the best-performing LLM (GPT-5.2).

\begin{table}[htbp]
\centering
\caption{Accuracy (\%) by question type}
\label{tab:accuracy}
\begin{tabular}{@{}lccccr@{}}
\toprule
\textbf{Question Type} & \textbf{\Mirror{}} & \textbf{GPT-5.2} & \textbf{GPT-5} & \textbf{Gemini-3-Pro} & \textbf{n\textsuperscript{\dag}} \\
\midrule
Diagnosis & 91.2 & 77.6 & 73.7 & 71.8 & 57 \\
Diagnostic Testing & 85.1 & 70.0 & 68.7 & 65.5 & 47 \\
Treatment & 86.3 & 70.8 & 72.2 & 64.4 & 73 \\
Risk/Prognosis & 86.8 & 80.0 & 81.6 & 71.1 & 38 \\
Pathophysiology & 86.7 & 74.7 & 73.8 & 76.1 & 60 \\
\midrule
\textbf{Overall} & \textbf{87.5} & \textbf{74.6} & \textbf{74.0} & \textbf{69.8} & \textbf{120} \\
\bottomrule
\end{tabular}

\vspace{0.5em}
{\small \textsuperscript{\dag}Questions may be tagged with multiple categories; category counts sum to $>$120.}
\end{table}

\begin{figure}[htbp]
\centering
\includegraphics[width=0.9\textwidth]{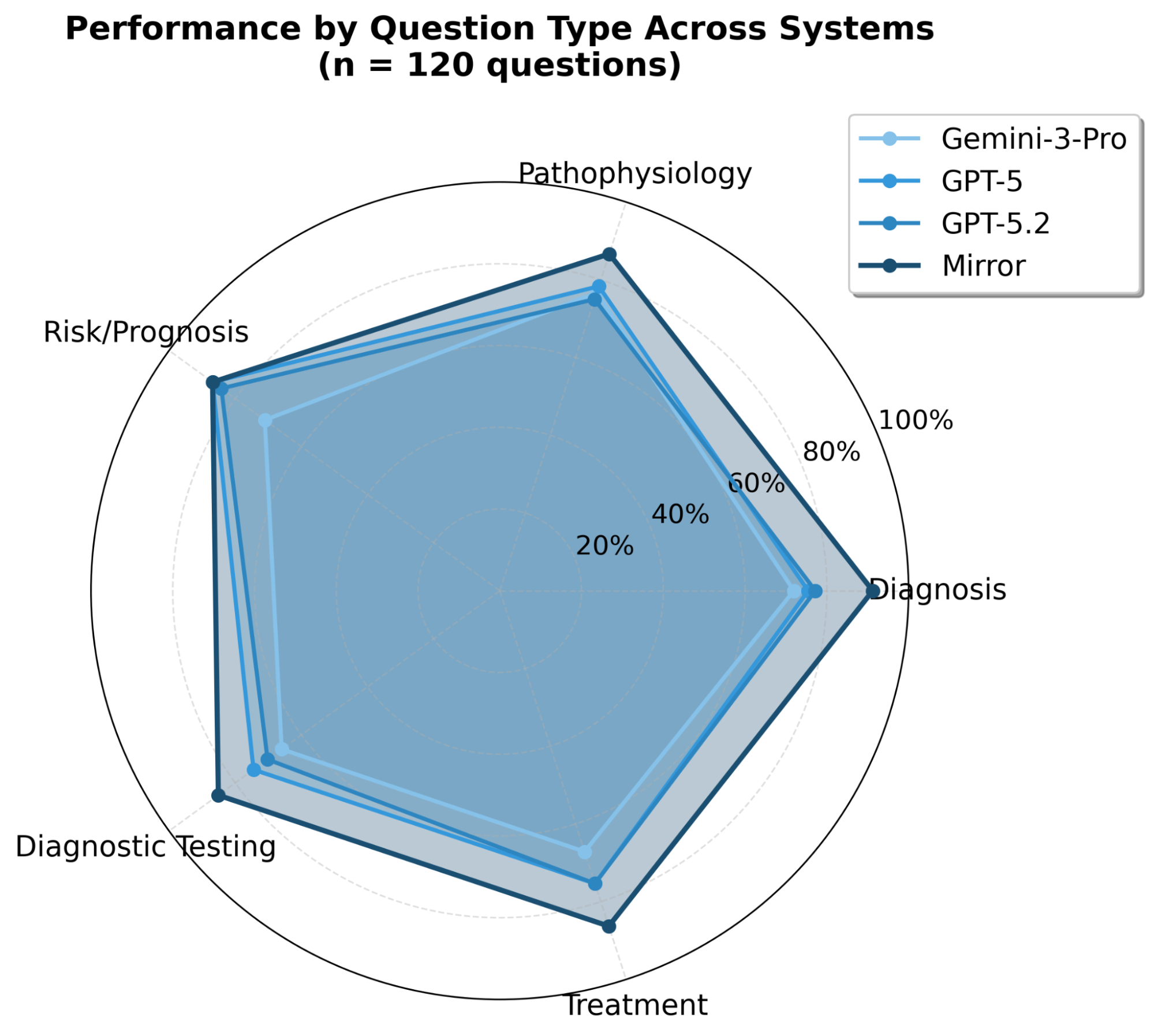}
\caption{Performance by question type across systems. \Mirror{} demonstrated consistent performance across all question categories, with the largest advantage in treatment questions (15.5 percentage point difference vs.\ GPT-5.2).}
\label{fig:questiontype}
\end{figure}

\begin{figure}[htbp]
\centering
\includegraphics[width=0.9\textwidth]{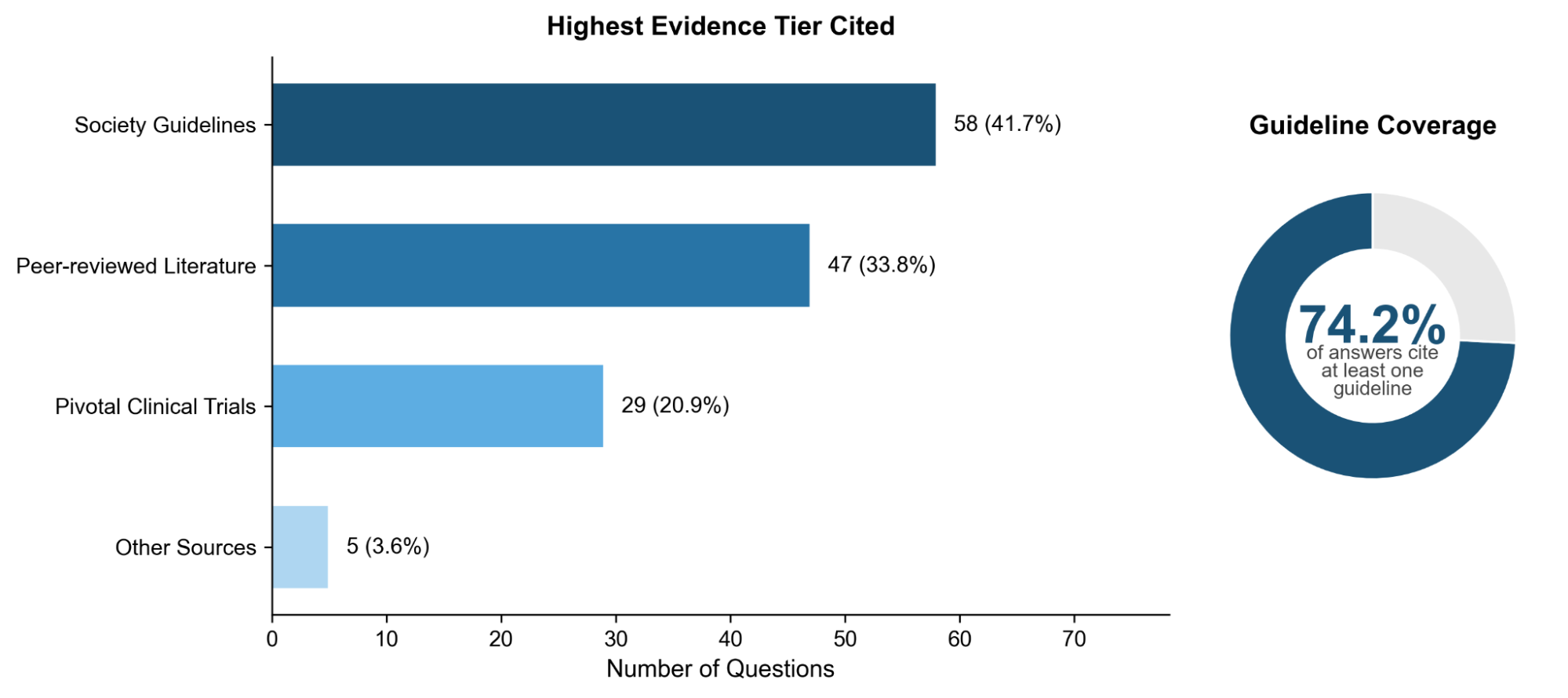}
\caption{Evidence provenance profile of \Mirror{} outputs. The left panel shows the highest evidence tier cited per question: society guidelines (41.7\%), peer-reviewed literature (33.8\%), pivotal clinical trials (20.9\%), and other sources (3.6\%). The right panel indicates that 74.2\% of answers included at least one guideline-tier citation, demonstrating the system's reliance on authoritative clinical guidance.}
\label{fig:provenance}
\end{figure}

\subsection{Error Analysis}

We conducted a qualitative analysis of questions answered incorrectly by each system to characterize failure modes. \Mirror{}'s errors ($n = 15$) were distributed across question types, with the highest concentration in treatment-tagged questions (10 of 15 errors involved treatment decisions). Error patterns included complex multi-domain questions requiring integration across diagnostic, pathophysiologic, and therapeutic reasoning, as well as cases involving nuanced interpretation of laboratory findings in atypical presentations. Representative examples are shown in Table~\ref{tab:errors}.

Frontier LLM errors showed a distinct pattern. GPT-5.2 frequently erred on questions requiring precise guideline-specific thresholds (e.g., A1c targets for specific patient populations, blood pressure goals in diabetic kidney disease) where the model's parametric knowledge conflicted with current recommendations. Gemini-3-Pro showed particular difficulty with questions involving multi-step clinical reasoning requiring integration of multiple evidence sources.

\begin{table}[htbp]
\centering
\caption{Representative error cases by system}
\label{tab:errors}
\small
\begin{tabular}{@{}p{2cm}p{6.5cm}p{5cm}@{}}
\toprule
\textbf{System} & \textbf{Clinical Scenario} & \textbf{Error Category} \\
\midrule
\Mirror{} & Pituitary macroadenoma with elevated prolactin and GH; selected dopamine agonist instead of confirming acromegaly & Incomplete biochemical evaluation before treatment \\
\addlinespace
\Mirror{} & Complete androgen insensitivity syndrome; prioritized bone density assessment over gonadal imaging & Metabolic workup prioritized over malignancy surveillance \\
\addlinespace
\Mirror{} & Opioid-induced hypogonadism with low gonadotropins; recommended obesity pharmacotherapy instead of pituitary imaging & Treatment prioritized over diagnostic workup for central hypogonadism \\
\addlinespace
\Mirror{} & Lactation-associated osteoporotic fracture with suboptimal calcium intake; recommended cessation of breastfeeding & Overaggressive intervention when conservative management indicated \\
\addlinespace
GPT-5.2 & Gestational diabetes with suboptimal control on basal insulin; selected outdated A1c target instead of current pregnancy-specific glucose thresholds & Guideline threshold: parametric knowledge conflicted with updated ADA recommendations \\
\addlinespace
Gemini-3-Pro & Type 2 diabetes with CKD and albuminuria on multiple agents; failed to prioritize SGLT2 inhibitor per KDIGO guidelines & Multi-step integration: did not synthesize cardiorenal benefit hierarchy across comorbidities \\
\bottomrule
\end{tabular}
\end{table}

To illustrate how reasoning patterns differ between \Mirror{} and frontier LLMs, Table~\ref{tab:comparison} presents side-by-side comparisons on two representative cases where \Mirror{} answered correctly and GPT-5.2 erred.

\begin{table}[htbp]
\centering
\caption{Side-by-side reasoning comparison: \Mirror{} vs GPT-5.2}
\label{tab:comparison}
\small
\begin{tabular}{@{}p{2.5cm}p{5.5cm}p{5.5cm}@{}}
\toprule
& \textbf{Case 1: Gestational Diabetes} & \textbf{Case 2: Acute Adrenal Hemorrhage} \\
\midrule
\textbf{Clinical Scenario} & Pregnant woman (14 weeks) with history of gestational diabetes and documented needle phobia causing insulin nonadherence in prior pregnancy, resulting in preeclampsia and preterm delivery. Now requires pharmacotherapy. & Previously healthy man presenting with acute abdominal pain, headache, and hypotension days after adenoviral-vector COVID-19 vaccine. CT shows bilateral adrenal hemorrhage. Labs show thrombocytopenia and markedly elevated D-dimer. \\
\addlinespace
\textbf{Correct Answer} & Metformin & IV hydrocortisone \\
\addlinespace
\textbf{\Mirror{} Answer} & Metformin \checkmark & IV hydrocortisone \checkmark \\
\addlinespace
\textbf{GPT-5.2 Answer} & Subcutaneous insulin & IV immunoglobulin \\
\addlinespace
\textbf{\Mirror{} Reasoning} & ``Treatment must be feasible and patient-specific. This patient has documented history of skipping insulin due to needle phobia, which already resulted in serious complications. Per ADA guidelines, metformin is an acceptable alternative when insulin is not feasible.'' & ``CT appearance consistent with acute adrenal hemorrhage. Patient is critically unwell with tachycardia and elevated lactate. Immediate priority is treating life-threatening adrenal insufficiency with IV hydrocortisone.'' \\
\addlinespace
\textbf{GPT-5.2 Reasoning} & ``Insulin is the preferred/standard pharmacotherapy in pregnancy because it does not cross the placenta. Address adherence by teaching partner to administer injections.'' & ``Presentation consistent with vaccine-induced immune thrombotic thrombocytopenia (VITT). Give IVIG to block Fc receptor-mediated platelet activation.'' \\
\addlinespace
\textbf{Failure Mode} & Rigid application of general guideline (``insulin is preferred'') without adapting to documented patient-specific barrier that caused prior treatment failure & Correctly identified underlying etiology (VITT) but prioritized disease-modifying therapy over immediate treatment of life-threatening adrenal crisis \\
\bottomrule
\end{tabular}
\end{table}

These cases illustrate two distinct failure patterns in frontier LLM clinical reasoning. In Case 1, GPT-5.2 correctly cited the general principle that insulin is preferred in pregnancy but failed to integrate the patient-specific context that made insulin adherence unlikely. \Mirror{}'s evidence corpus includes ADA guidance explicitly addressing scenarios where insulin is not feasible, enabling appropriate patient-centered adaptation. In Case 2, GPT-5.2 demonstrated sophisticated pattern recognition in identifying VITT but applied a disease-focused rather than patient-focused management hierarchy. \Mirror{} correctly prioritized the immediate life-threatening complication (adrenal insufficiency) over treatment of the underlying thrombotic process.

\subsection{Evidence Traceability and Citation Accuracy}

\Mirror{} produced evidence-linked outputs for all 120 questions, with each answer accompanied by citations to specific sources in the curated corpus. Analysis of evidence provenance (Figure~\ref{fig:provenance}) revealed that 41.7\% of outputs cited guideline-tier sources as the highest evidence tier, with 74.2\% of answers including at least one society guideline citation.

Manual verification of a random sample of 60 questions demonstrated 100\% citation accuracy (60/60 verified), with all sampled citations meeting all three verification criteria: the cited source existed in the corpus, the passage was accurately represented, and the evidence logically supported the selected answer. Verification was performed by a clinical reviewer who was not involved in system development.

For treatment-related questions ($n = 73$), \Mirror{} achieved 86.3\% guideline concordance, with 63 of 73 answers matching the relevant society guideline recommendation. This high concordance rate reflects \Mirror{}'s systematic prioritization of authoritative clinical guidance in therapeutic decision-making.

\section{Discussion}

This study demonstrates that domain-specific evidence curation and structured clinical reasoning can achieve subspecialty-level performance exceeding both frontier LLMs and a human reference from practicing endocrinologists. \Mirror{}'s 87.5\% accuracy on the ESAP 2025 endocrinology examination represents a 25.2 percentage point improvement over the human reference (62.3\%) and a 12.9 percentage point improvement over the best-performing general-purpose LLM and approaches an upper-bound reference condition in which a baseline model is provided with manually selected, question-specific evidence passages. Notably, \Mirror{}'s advantage was most pronounced on difficult questions where human accuracy fell below 50\%, suggesting the system captures clinical reasoning patterns that challenge even subspecialty-trained physicians.

\subsection{Why Curated Evidence Outperforms Web Retrieval}

The performance differential can be attributed to three advantages of curated evidence infrastructure over unconstrained web retrieval.

\textbf{First, evidence curation provides quality control that web retrieval cannot match.} \Mirror{}'s corpus includes only peer-reviewed guidelines and high-impact clinical literature, with explicit tiering by evidence strength. Web retrieval, by contrast, can surface heterogeneous sources with variable provenance and without a unified evidence-grading layer, which makes it difficult to consistently prioritize guideline-grade recommendations over lower-signal material in a repeatable way. Recent systematic reviews have highlighted that retrieval-augmented generation (RAG) can reduce hallucination rates in healthcare applications \citep{amugongo2025}, but the quality of the underlying corpus remains critical. This matters particularly for clinical reasoning, where the strength of underlying evidence should inform confidence in recommendations.

\textbf{Second, structured evidence retrieval enables clinical relevance that semantic similarity cannot capture.} Clinical questions often hinge on specific thresholds, contraindications, or guideline-specific recommendations that may not be semantically similar to query terms. A question about blood pressure targets in diabetic nephropathy requires retrieval of KDIGO guidelines, not just any content mentioning blood pressure and diabetes. \Mirror{}'s domain-specific retrieval is tuned for these clinical relevance patterns.

\textbf{Third, curated retrieval enables complete traceability.} Every \Mirror{} output can be linked to specific, verifiable evidence sources, with 100\% citation accuracy upon manual verification. This auditability is not achievable with web retrieval, where source quality varies and provenance may be unclear. For clinical deployment, traceability is not merely desirable but essential, clinicians must be able to verify AI recommendations against primary evidence.

\subsection{Clinical Implications}

These findings have direct implications for the development of clinical decision support systems. The dominant paradigm, larger models with web access, may be insufficient for subspecialty applications where reliability is paramount. Instead, our results suggest that investment in domain-specific evidence infrastructure may yield greater returns than model scale for clinical AI.

The top-2 accuracy of 92.5\% is particularly relevant for clinical workflows. Decision support systems need not provide single definitive answers; rather, they can serve as cognitive aids that help clinicians efficiently identify the most probable diagnoses or appropriate treatment options. \Mirror{}'s high top-2 accuracy suggests utility in this workflow even when single-answer accuracy is imperfect.

Furthermore, \Mirror{}'s architecture provides inherent traceability, every output can be linked to specific evidence sources, enabling clinicians to verify recommendations against primary literature. This auditability addresses a key barrier to clinical AI adoption: the ``black box'' concern that LLM outputs cannot be validated \citep{lee2023}.

\subsection{Limitations}

Several limitations should be considered when interpreting these results. First, this evaluation focused on a single subspecialty (endocrinology); generalization to other domains requires further validation. We selected endocrinology because of its clinical importance and examination rigor, but performance in cardiology, nephrology, or other subspecialties may differ.

Second, board-style examinations, while rigorous, differ from real clinical practice. ESAP questions present well-defined vignettes with single correct answers; actual clinical encounters involve ambiguity, incomplete information, and patient preferences that are not captured in standardized testing. Our ongoing work includes evaluation on real-world clinical scenarios with human clinician assessment.

Third, the human reference accuracy (62.3\%) is derived from aggregate ESAP respondent statistics rather than individual test-taker scores, representing the mean percentage of respondents selecting the correct answer per question. ESAP respondents are self-selected endocrinologists pursuing certification maintenance, and testing conditions differ from AI evaluation (time pressure, fatigue, no external resources). This reference provides useful context but should not be interpreted as a direct head-to-head comparison with typical clinical practice.

Fourth, the comparison between \Mirror{} (curated corpus, no web) and frontier LLMs (web retrieval) confounds multiple variables: corpus quality, retrieval method, and access modality. \Mirror{}'s superior performance may reflect advantages of evidence curation, but could also reflect that the curated corpus happened to contain content well-suited to ESAP questions. Future work comparing systems under matched retrieval conditions would help isolate the contribution of corpus curation versus retrieval architecture.

Fifth, we evaluated a limited set of baseline systems and configurations; results may differ for other clinical AI products, alternative prompting/browsing protocols, or future model releases.

Sixth, ESAP 2025 is a proprietary assessment and the original question text cannot be shared publicly, which limits direct reproducibility.

Finally, both frontier model behavior and web retrieval toolchains can change over time (``tool drift''), which may affect absolute baseline performance; we therefore report evaluation dates and protocols to support fair longitudinal interpretation. The evidence corpus is currently U.S.-centric, reflecting American society guidelines; application in other healthcare systems with different standards would require corpus adaptation.

\subsection{Future Directions}

This work establishes a foundation for several directions of ongoing investigation. We are extending the benchmark evaluation to additional subspecialties, including cardiology, nephrology, and general internal medicine, to assess generalization of the evidence-grounded approach. We are also conducting prospective evaluation with subspecialist clinicians using real-world case vignettes to assess clinical utility beyond examination accuracy. These studies will address whether board examination performance translates to meaningful decision support in practice.

\section{Conclusions}

\Mirror{}, an evidence-grounded clinical reasoning system, achieved 87.5\% accuracy on a subspecialty endocrinology board examination, exceeding frontier LLMs by 12.9 percentage points despite those models having web access that \Mirror{} lacked. Critically, \Mirror{} provided complete evidence traceability, 74.2\% of outputs cited at least one guideline-tier source with 100\% verified citation accuracy, enabling clinicians to verify every recommendation against primary evidence. These results demonstrate that curated, high-quality evidence bases with explicit provenance can outperform unconstrained web retrieval for subspecialty clinical reasoning while simultaneously addressing the auditability requirements essential for clinical deployment.

\section*{Data Availability}

The ESAP 2025 questions are copyrighted by the Endocrine Society and cannot be redistributed; evaluation materials are therefore not publicly released. Code for statistical analyses is available upon request.

\section*{Conflicts of Interest}

Authors A.H., R.Z., U.M., and G.S.\ are employees of January AI, which developed \Mirror{}. Authors K.K.\ and N.A.\ serve as advisors to January AI.

\bibliographystyle{unsrtnat}

\end{document}